\documentclass{article}
\usepackage{spconf,amsmath,graphicx}
\usepackage{graphics}
\usepackage{amssymb}
\usepackage{mathrsfs}
\usepackage{epstopdf}
\usepackage{algorithmic}
\usepackage{algorithm}
\usepackage{subfigure}
\usepackage{multicol}
\usepackage{multirow}
\usepackage{url}

\def\ie{\emph{i.e.}}

\title{Multiple Instance Hybrid Estimator for Learning Target Signatures}
\name{Changzhe Jiao and Alina Zare\thanks{This material is based upon work supported by the National Science Foundation under Grant IIS-1350078-CAREER: Supervised Learning for Incomplete and Uncertain Data and a scholarship from China Scholarship Council (No. 201206960005).}}
\address{ Electrical and Computer Engineering, University of Missouri\\ Electrical and Computer Engineering, University of Florida}
\begin{document}
	%
	\maketitle
	\begin{abstract}
	Signature-based detectors for hyperspectral target detection rely on knowing the specific target signature in advance. However, target signature are often difficult or impossible to obtain. Furthermore, common methods for obtaining target signatures, such as from laboratory measurements or manual selection from an image scene, usually do not capture the discriminative features of target class. In this paper, an approach for estimating a discriminative target signature from imprecise labels is presented. The proposed approach maximizes the response of the hybrid sub-pixel detector within a multiple instance learning framework and estimates a set of discriminative target signatures. After learning target signatures, any signature based detector can then be applied on test data. Both simulated and real hyperspectral target detection experiments are shown to illustrate the effectiveness of the method. 
		
	\end{abstract}
	\begin{keywords}
		target detection, concept learning, hyperspectral, multiple instance, target characterization, subpixel.
	\end{keywords}
	\section{Introduction}
	\label{sec:intro}
	The wealth of spectral information in hyperspectral imagery provides for the ability to perform sub-pixel target detection.   Signature-based hyperspectral target detectors need to know the target signature in advance. However, in a number of scenarios, obtaining an effective target signature is often a challenging problem.  For instance, in some situations analysts may have approximate locations of sub-pixel targets and want to estimate their target signature and search for the target material elsewhere. In these situations, manual selection of a target signature from an image cube would not only be difficult (as only approximate locations are known) but may result in selecting a mixed (as opposed to pure target) pixels. 
	
	In this paper, we address this problem by proposing a method to characterize a target signature from imprecisely labeled hyperspectral imagery.  Specifically, we model the hyperspectral target estimation task as a multiple instance concept learning problem. In multiple instance learning (MIL)  \cite{Dietterich:1997}, training data is partitioned into sets of labeled bags (instead of being individually labeled). A positive bag must contain at least one true positive (target) data point and negative bags are composed entirely of negative data. Multiple instance concept learning is a branch of MIL that aims to learn one or a set of concepts to describe the target class. For example, Diversity Density (DD) \cite{Maron:1998} identifies a target concept that is close to the intersection of all positive bags and far from negative instances. The expectation maximization version of DD \cite{Zhang:2002} improves  convergence. Previous methods for MIL target characterization were proposed in \cite{Zare:2015fumi, zare2016miace}.  \cite{Zare:2015fumi} combines all positive bags into one large positive bag and, thus, discards bag-level label information.  \cite{zare2016miace} does not assume a mixing model and, thus, may fail to take advantage of the mixing model if it is known. In contrast, the proposed method maximizes the response of the hybrid sub-pixel detector, preserves bag-level labels, and assumes the linear mixing model \cite{Broadwater:2007}.

	\section{Multiple Instance Hybrid Estimator}
	\label{sec:MIHE}
	
	Let $\mathbf{X}=\left[\mathbf{x}_1,\cdots,\mathbf{x}_N\right]\in\mathbb{R}^{d\times N}$ be training data where $d$ is the dimensionality of an instance and $N$ is the total number of training instances. The data is grouped into $K$ \textit{bags},  $\mathbf{B} = \left\{ \mathbf{B}_1, \ldots, \mathbf{B}_K\right\}$, with associated binary bag-level labels, $L = \left\{L_1, \ldots, L_K\right\}$ where $L_i \in \left\{ 0, 1\right\}$; $n_i$ is the number of instances in bag $\mathbf{B}_i$ and $\mathbf{x}_{ij} \in \mathbf{B}_i$ denotes the $j^{th}$ instance in bag $\mathbf{B}_i$ with instance-level label $l_{ij}\in\left\{ 0, 1\right\}$. When identifying the label on certain bag or instance is important,  positive bags will be indicated as $\mathbf{B}_i^+$ with associated bag level label $L_i^+$ containing instances $\mathbf{x}_{ij}^+$ with instance-level labels $l_{ij}^+, j=1,\cdots, n_i^+$, where $n_i^+$ is the number of instances in positive bag $\mathbf{B}_i^+$. Similarly, $\mathbf{B}_i^-,\: L_i^-,\: \mathbf{x}_{ij}^-,\: l_{ij}^-$ and $n_i^-$ represent a negative bag, label for this negative bag, the $j$th instance in this bag, label for the $j$th instance and total number of instances in this bag.  The number of positive and negative bags are denoted as $K^+$ and $K^-$, $N^+$ and $N^-$ represent the total number of positive and negative instances, respectively. Thus $N=N^++N^-=\sum_{i=1}^{K^+}n_i^++\sum_{i=1}^{K^-}n_i^-$ and $\mathbf{B}^+$,  $\mathbf{B}^-$ represent the union of all positive instances and negative instances, respectively.
	
	Given this notation, the proposed multiple instance hybrid estimator (MI-HE) aims to maximize the probability of the labels of the  bags as in \eqref{eq:MIHD_bag_likelihood}.  Since, in MIL, each positive bag must contain at least one positive instance, we can then substitute the probability for a bag to be positive with the probability of the most likely point in the bag to be a target, \ie, $\max_{j\in n_i^+}\Pr(\mathbf{x}_{ij}^+=+|\mathbf{B}_i^+)$.  Since each negative bag is assumed to only contain non-target instances, then the probability for a negative bag to be negative can be represented by the joint probability of all instances in this bag to be negative,
	\begin{small}
	\begin{eqnarray}
	J_1&=&\prod_{i=1}^{K^+} \Pr(L_i^+=+|\mathbf{B}_i^+)\prod_{i=1}^{K^-}\Pr(L_i^-=-|\mathbf{B}_i^-).
	\label{eq:MIHD_bag_likelihood}\\
	&=&\prod_{i=1}^{K^+} \max_{j\in n_i^+}\Pr({l}_{ij}^+=+|\mathbf{B}_i^+) \prod_{i=1}^{K^-}\prod_{j=1}^{n_i^-}\Pr({l}_{ij}^-=-|\mathbf{x}_{ij}^-).
	\label{eq:MIHD_inst_likelihood}
	\end{eqnarray}
	\end{small}

	Eq. \eqref{eq:MIHD_inst_likelihood} contains a $\max$ operation that is difficult to optimize numerically. Some algorithms in the literature adopt a noisy-OR model instead of a $\max$ \cite{Maron:1998, Zhang:2002}. However, experimental results show that the noisy-OR is non-smooth and generally needs to be optimized repeatedly with many different initializations to avoid local optima. In the proposed approach, we adopt the generalized mean as an alternative,
	\begin{small}
	\begin{equation}
	J_2=\prod_{i=1}^{K^+} \left(\frac{1}{n_i^+}\sum_{j=1}^{n_i^+}\Pr({l}_{ij}^+=+|\mathbf{B}_i^+)^p\right)^{\frac{1}{p}} \prod_{i=1}^{K^-}\prod_{j=1}^{n_i^-}\Pr({l}_{ij}^-=-|\mathbf{x}_{ij}^-),
	\label{eq:MIHD_gen_mean}
	\end{equation}
	\end{small}
	where $p\in [-\infty, +\infty]$ is a parameter that varies the operation from a $\min$ to a $\max$, respectively. We then optimize the negative logarithm of $J_2$ and add a scaling factor, $\rho < 1$, to the second term to control the influence of the negative bags (when $N^- \ne N^+$) as shown in \eqref{eq:MIHD_neg_log}.
\begin{figure*}[!t]
	\begin{equation}
	-\ln J=-\sum_{i=1}^{K^+}\frac{1}{p}\ln \left(\frac{1}{n_i^+}\sum_{j=1}^{n_i^+}\Pr({l}_{ij}^+=+|\mathbf{B}_i^+)^p\right)-\rho \sum_{i=1}^{K^-}\sum_{j=1}^{n_i^-}\ln\Pr({l}_{ij}^-=-|\mathbf{x}_{ij}^-),
	\label{eq:MIHD_neg_log}
	\end{equation}
	\end{figure*}
	
	We now must define $\Pr({l}_{ij}^+=+|\mathbf{B}_i^+)$ and $\Pr({l}_{ij}^-=-|\mathbf{x}_{ij}^-)$. As in \cite{jiao2016ICPR}, each instance is modeled as a sparse linear combination of target and/or background signatures $\mathbf{D}$, $\mathbf{x}_j\approx\mathbf{D}\boldsymbol{\alpha}_j$, where $\boldsymbol{\alpha}_j$ is the sparse vector of  weights. Each positive bag contains at least one instance with target:
	\begin{eqnarray}
	&&\text{if }L_i = 1,  \nonumber \exists \mathbf{x}_j \in \mathbf{B}_i^+ \text{ s.t. } \\
	&&\mathbf{x}_j = \sum_{t=1}^{T}\alpha_{jt}\mathbf{d}_t^+ + \sum_{k=1}^{M} \alpha_{jk}\mathbf{d}_{k}^-+\boldsymbol{\varepsilon}_{j}, \alpha_{jt} \ne 0,
	\label{eq:MIHD_model_pos}
	\end{eqnarray}
	where $\boldsymbol{\varepsilon}_j$ is a noise term. A negatively labeled bag $\mathbf{B}_i^-$ should not contain any target:
	\begin{equation}
	\text{if }L_i = 0,  \forall \mathbf{x}_j \in \mathbf{B}_i^-, \mathbf{x}_j =  \sum_{k=1}^{M} \alpha_{jk}\mathbf{d}_{k}^-+\boldsymbol{\varepsilon}_{j}.
	\label{eq:MIHD_model_neg}
	\end{equation}
	
	Given this model, we introduce the hybrid subpixel detector to estimate the probability instances from positive bags are positive. Specifically, the probability for $\mathbf{x}_{ij}^+$ in $\mathbf{B}_i^+$ is a target point is defined as: 
	\begin{equation}
	\Pr(l_{ij}^+=+|\mathbf{B}_i^+)=\exp\left(-\beta\frac{\|\mathbf{x}_{ij}^+-\mathbf{D}\boldsymbol{\alpha}_{ij}^+\|^2}{\|\mathbf{x}_{ij}^+-\mathbf{D}^-\boldsymbol{\alpha}_{ij}^{+b}\|^2}\right),
	\label{eq:MIHD_pos_inst_model}
	\end{equation}
	where $\mathbf{D}=\begin{bmatrix}\mathbf{D}^+ & \mathbf{D}^-\end{bmatrix}\in\mathbb{R}^{d\times (T+M)}$, $\mathbf{D}^+ = \left[\mathbf{d}_{1}^+,\cdots,\mathbf{d}_{T}^+\right]$ is the set of $T$ target  signatures and $\mathbf{D}^- = \left[\mathbf{d}_{1}^-,\cdots,\mathbf{d}_{M}^-\right]$ is the set of $M$ background  signatures, $\beta$ is a scaling parameter;  $\boldsymbol{\alpha}_{ij}^{+}$ and $\boldsymbol{\alpha}_{ij}^{+b}$ are the sparse representation of $\mathbf{x}_{ij}^+$ given entire signatures set $\mathbf{D}$ and background signatures set $\mathbf{D}^-$, respectively. Specifically, solving for a sparse $\boldsymbol{\alpha}$ given a dictionary set $\mathbf{D}$ is modeled as the Lasso problem \cite{tibshirani1996regression, chen2001atomic} shown in \eqref{eq:MIHD_lasso}:
	\begin{equation}
	\hat{\boldsymbol{\alpha}}=\arg\min\frac{1}{2}\|\mathbf{x}-\mathbf{D}\boldsymbol{\alpha}\|^2_2+\lambda\|\boldsymbol{\alpha}\|_1,
	\label{eq:MIHD_lasso}
	\end{equation}
	where $\lambda$ is a scaling vector to control the sparsity of $\boldsymbol{\alpha}$. Here we adopt the iterative shrinkage-thresholding algorithm (ISTA) \cite{daubechies2003iterative} for solving the sparse codes  $\boldsymbol{\alpha}$.
	
		\begin{algorithm}
			\caption{MI-HE algorithm}
			\algsetup{indent=1em}
			\begin{algorithmic}[1] 
				\STATE Initialize $\mathbf{D}^0$, $iter = 0$
				\REPEAT
				\FOR{$t=1,\cdots,T$}
				\STATE Solving $\begin{small}\left\{\boldsymbol{\alpha}^+_i \right\}_{i=1}^{N}\end{small}$, $\begin{small}\left\{\boldsymbol{\alpha}^-_i \right\}_{i=1}^{N}\end{small}$ by ISTA
				\STATE Update  $\mathbf{d}_t$ by optimizing \eqref{eq:MIHD_neg_log} using gradient descent
				\STATE $\mathbf{d}_t\gets\frac{1}{\|\mathbf{d}_t\|_2}\mathbf{d}_t$
				\ENDFOR
				\FOR{$k=1,\cdots,M$}
				\STATE Solving  $\begin{small}\left\{\boldsymbol{\alpha}^+_i \right\}_{i=1}^{N}\end{small}$, $\begin{small}\left\{\boldsymbol{\alpha}^-_i \right\}_{i=1}^{N}\end{small}$ by ISTA
				\STATE Update  $\mathbf{d}_k$ by optimizing \eqref{eq:MIHD_neg_log} using gradient descent
				\STATE $\mathbf{d}_k\gets\frac{1}{\|\mathbf{d}_k\|_2}\mathbf{d}_k$
				\ENDFOR
				\STATE $iter \gets iter + 1$
				\UNTIL{Stopping criterion meets}
				\RETURN $\mathbf{D}$\\
			\end{algorithmic} 
			\label{alg:MIHD}
		\end{algorithm}
		
	For points from negative bags, following \eqref{eq:MIHD_model_neg}, we model the reconstruction error of points $\mathbf{x}_{ij}^- \in \mathbf{B}_i^+$ as a zero mean Gaussian distribution with unknown variance,
	\begin{equation}
	\Pr({l}_{ij}^-=-|\mathbf{x}_{ij}^-)=\exp\left(\|\mathbf{x}_{ij}^--\mathbf{D}^-\boldsymbol{\alpha}_{ij}^-\|^2\right),
	\label{eq:MIHD_neg_inst_model}
	\end{equation}
	where $\boldsymbol{\alpha}_{ij}^-$ is the sparse representation of $\mathbf{x}_{ij}^-$ given $\mathbf{D}^-$.  The objective function \eqref{eq:MIHD_neg_log} is then optimized by gradient descent with sparse coding as outlined in Alg. \ref{alg:MIHD}

	\section{EXPERIMENTS}
	\label{sec:experiments}
	
\begin{figure}
		\centering
		\includegraphics[width=5cm]{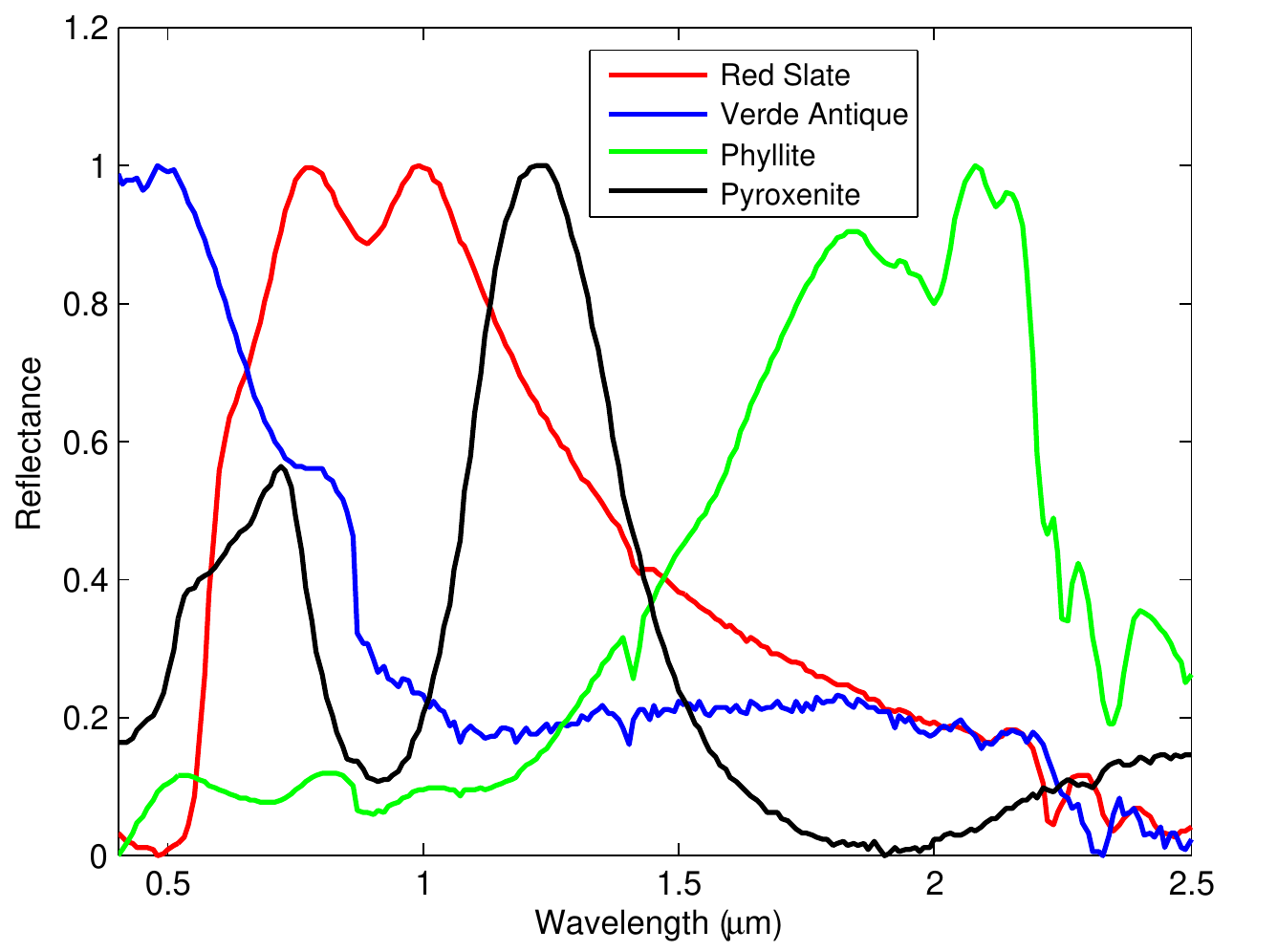}
		\caption{Signatures used to generate simulated data \label{fig:constituent_endmembers} } 
	\end{figure}

	MI-HE was applied to simulated data generated from four spectra selected from the ASTER spectral library \cite{aster:2009} shown in Fig. \ref{fig:constituent_endmembers}. Specifically, the Red Slate, Verde Antique, Phyllite and Pyroxenite spectra from the rock class with 211 bands and wavelengths ranging from $0.4 \mu$m to $2.5 \mu$m were used as endmembers to generate hyperspectral data. Red Slate was labeled as the target endmember. \cite{Zare:2015fumi} provides a precise description of how the simulated data was generated. 
	
		\begin{table} 
			\begin{footnotesize}
				\begin{center}
					\caption{List of Background Endmembers for Synthetic Data}\label{tab:toydata_endmember_list}
					\begin{tabular}{|c|c|c|}
						\hline
						Bag Number 	&  Bag Label  & Background Endmember \\
						\hline\hline
						1-5 &     $+$     &        Verde Antique, Phyllite, Pyroxenite    \\\hline
						6-10&          $+$        &     Phyllite, Pyroxenite                       \\\hline
						11-15&      $+$         &      Pyroxenite                    \\\hline
						16-20&       $-$        &      Phyllite, Pyroxenite                 \\\hline
					\end{tabular}
				\end{center}
			\end{footnotesize}
		\end{table}
		
	Three sets of highly-mixed noisy data with varied mean target proportion value ($p_{t\_mean}$) were generated. Specifically, this synthetic data has 15 positive and 5 negative bags with each bag has 500 points. If it is positively labeled, there are 100 target points containing mean target (Red Slate) proportion $p_{t\_mean}$. The parameter $p_{t\_mean}$ that controls the mean target proportion value was set to 0.3, 0.5 and 0.7 respectively to vary the level of target presence from weak to high. Gaussian white noise was added so that signal-to-noise ratio of the data was set to $30 dB$. To highlight the ability of MI-HE to leverage individual  bag-level labels, we use different subsets of background endmembers to build synthetic data as shown in Tab. \ref{tab:toydata_endmember_list}.

\begin{figure}
		\begin{center}
			\subfigure[Estimated target spectra]{   
				\includegraphics[width=4cm]{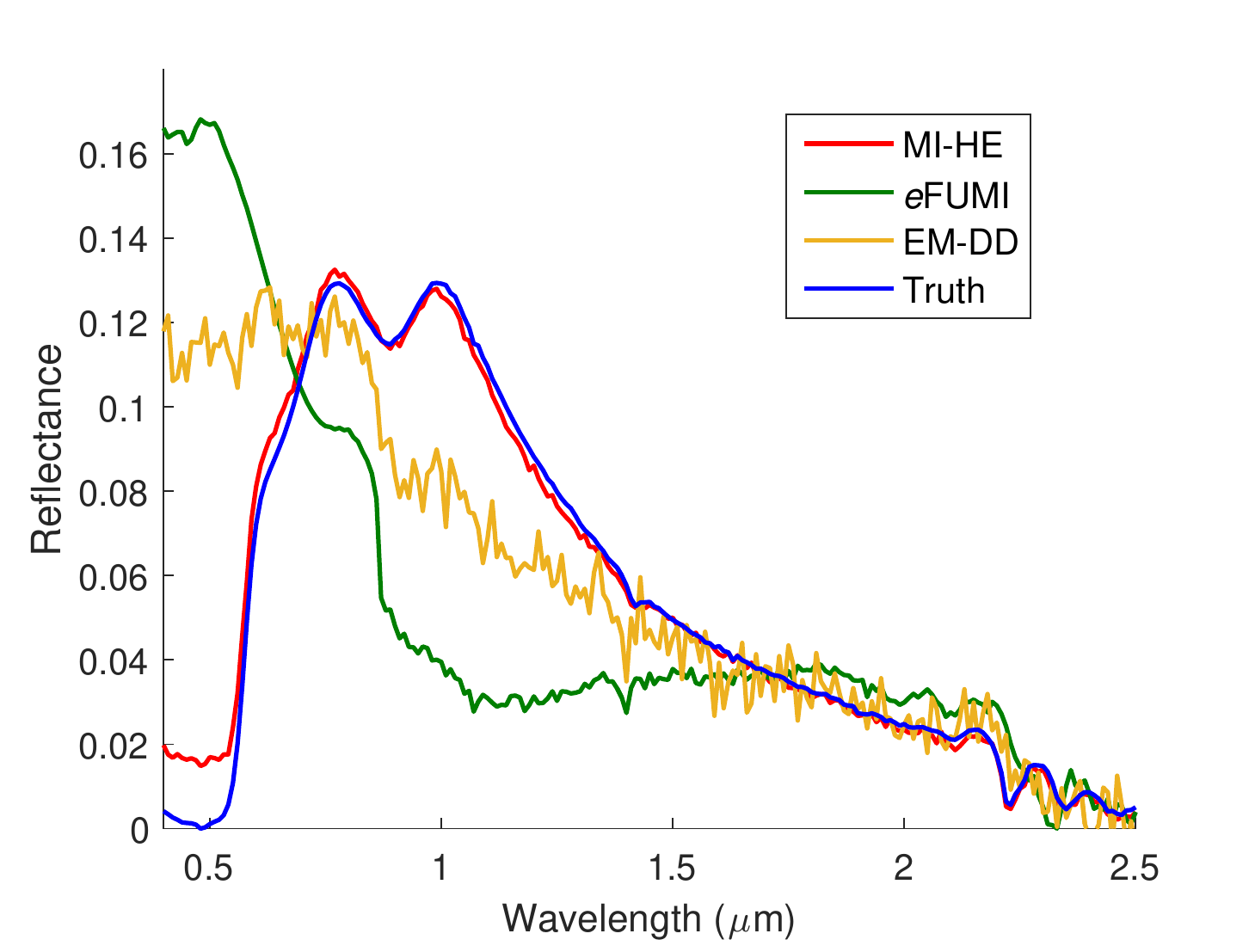} \label{fig:sig_plot_toydata_ptmean03}}
			\subfigure[ROC curves]{   
				\includegraphics[width=4cm]{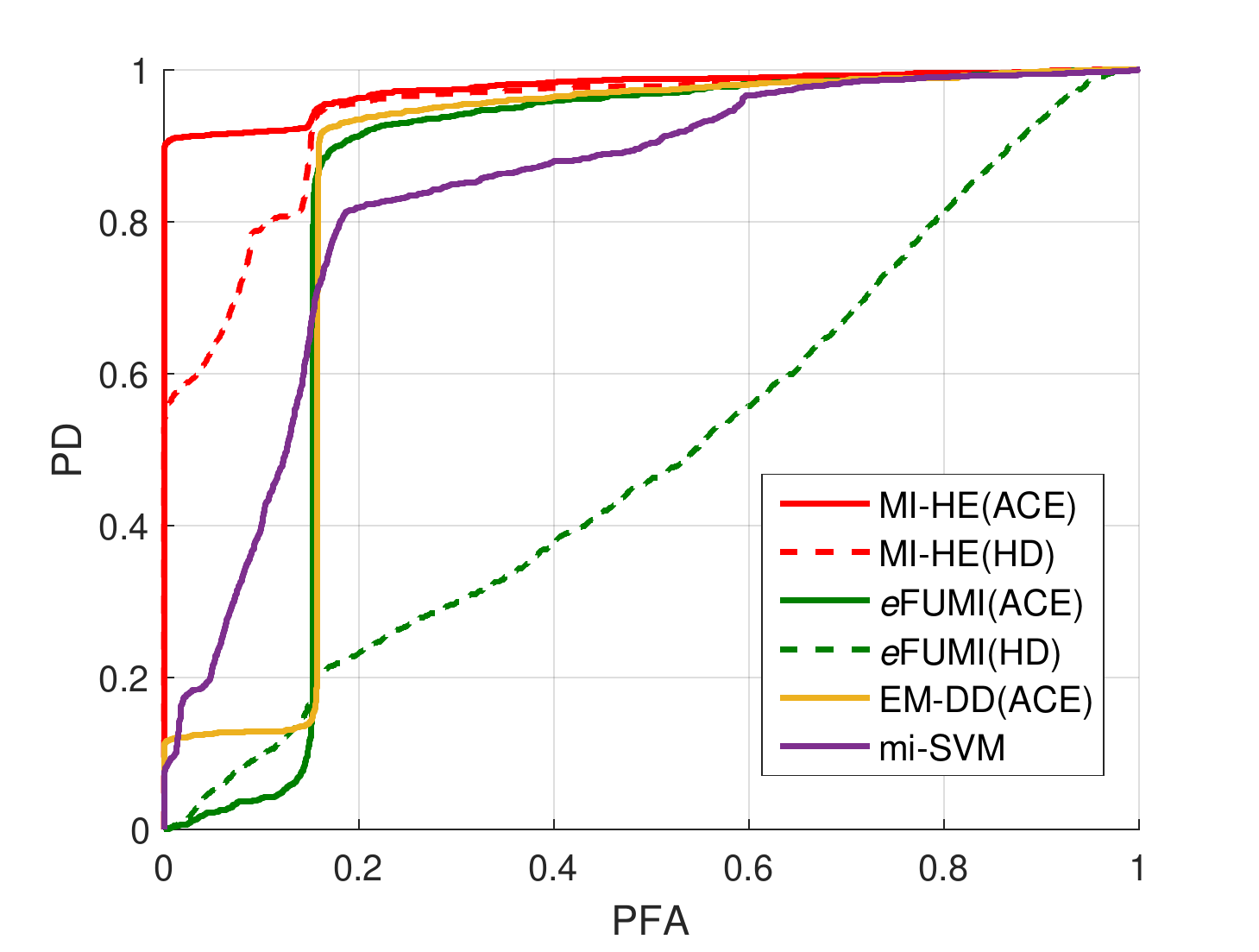} \label{fig:rocs_toydata_ptmean03}}
			\caption{Comparison of MI-HE, $e$FUMI, EM-DD and mi-SVM on synthetic data with mean target proportion 0.3}\label{fig:compare_tyodata_ptmean_03}
		\end{center}
	\end{figure}

	MI-HE was compared to $e$FUMI and EM-DD. As we mentioned before, $e$FUMI combines all positive bags into one big positive bag. Given the aforementioned synthetic data, $e$FUMI confuses both Red Slate and Verde Antique as target signatures since Verde Antique is missing in the training negative bags (and $e$FUMI does not preserve bag-level labels). However, the proposed MI-HE is able to learn the target signature correctly. Fig. \ref{fig:sig_plot_toydata_ptmean03} shows the target signature estimated by MI-HE and comparison algorithms $e$FUMI \cite{Zare:2015fumi} and EM-DD \cite{Zhang:2002}. We can see that MI-HE is able to learn a target signature that matches the groundtruth, but $e$FUMI mistakes Verde Antique as target signature and EM-DD learns one noisy, non-pure target signature.

	For quantitative evaluation, receiver operating curves (ROC) from detection on testing data generated using the same method are shown in Fig. \ref{fig:rocs_toydata_ptmean03}. Tab. \ref{tab:AUC_toydata} shows the area of probability of detection (PD) under the curve (AUC), where we can see proposed MI-HE outperforms the comparison algorithms $e$FUMI, EM-DD and mi-SVM \cite{andrews:2002}. For MI-HE and $e$FUMI, both the structured-background adaptive coherence/cosine estimator (ACE) \cite{Kraut:1999, basener:2010clutter} and Hybrid Sub-pixel Detector (HD) \cite{Broadwater:2007} were applied; for EM-DD, only ACE was applied since EM-DD does not learn a set of background signatures simultaneously. The results reported are the median results over five runs of the algorithm on the same data.

\begin{table} 
	\begin{center}
		\caption{Simulated Hyperspectral Data Experiments. Results listed in Median AUC over 5 Runs}\label{tab:AUC_toydata}
		\begin{tabular}{|c|c|c|c|}
			\hline
			\multirow{2}{*}{Algorithm} 	&  \multicolumn{3}{c|}{$P_{t\_mean}$} \\
			\cline{2-4}&{0.3}&{0.5}&{0.7}\\
			\hline\hline
			{MI-HE (HD)}      &   \underline{0.944}      &      \underline{0.981}              &   0.996     \\\hline
			{MI-HE (ACE)}     &      \textbf{0.974}      &      \textbf{0.997}     &   \textbf{0.999}    \\\hline
			{$e$FUMI (ACE)}  &      0.810                 &      0.843                &   0.843    \\\hline
			{$e$FUMI (HD)}   &      0.471                 &      0.467                &   0.504     \\\hline
			{EM-DD (ACE)}    &      0.841               &      {0.876}  &   \underline{0.998}    \\\hline
			{mi-SVM}         &      0.834               &      0.884              &   {0.891}    \\\hline
		\end{tabular}
	\end{center}
\end{table}

\begin{figure}
	\centering
	{\includegraphics[width=7cm]{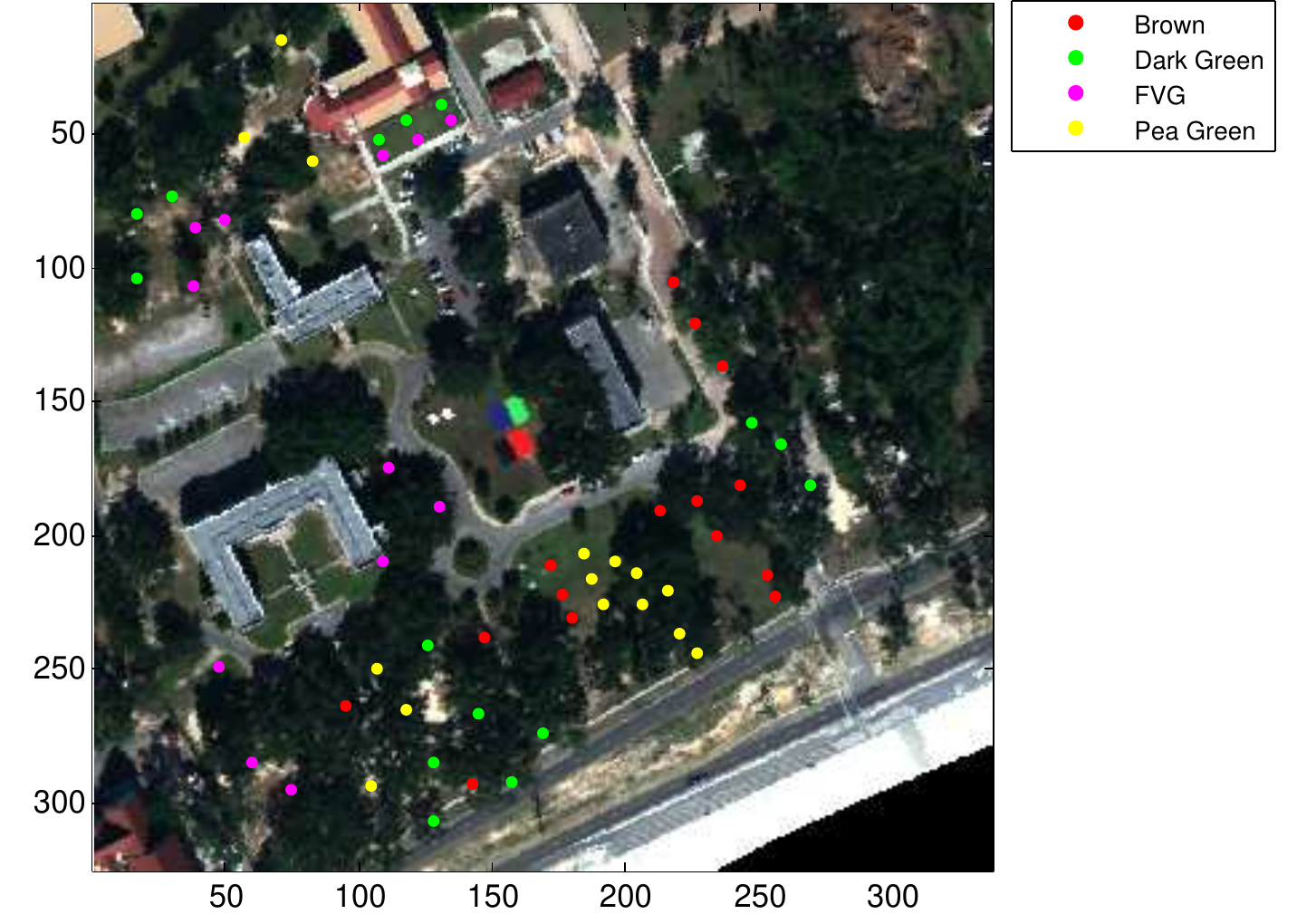}}
	\caption{MUUFL Gulfport data set RGB image and the 64 target locations}
	\label{fig:gulfport_rgb}
\end{figure}

\begin{table} 
	\begin{scriptsize}
		\begin{center}
			\caption{Gulfport Brown Detection. Results listed in Median NAUC over 5 Runs at FAR $=1\times10^{-3} m^2$}\label{tab:AUC_gulfport}
			\begin{tabular}{|c|c|c|}
				\hline
				{Algorithm} 	&  {Train: Flight 1, Test: Flight 3}& {Train: Flight 3, Test: Flight 1} \\
				\hline\hline
				{MI-HE (HD)}&      \textbf{0.552}     &  \textbf{0.792}         \\\hline
				{MI-HE (ACE)}&          \underline{0.442}         &     {0.699}                       \\\hline
				{$e$FUMI (ACE)}&       0.437         &      0.746                    \\\hline
				{$e$FUMI (HD)}&        0.417         &      \underline{0.765}                     \\\hline
				{EM-DD (ACE)}&        {0.420}   &     {0.749}                      \\\hline
				{mi-SVM}&        0.353         &   0.333      \\\hline
			\end{tabular}
		\end{center}
	\end{scriptsize}
\end{table}

  For experiments on real hyperspectral target detection data, the MUUFL Gulfport Hyperspectral data set was used.  This data set was collected over the University of Southern Mississippi-Gulfpark Campus and contains $325\times337$ pixels with 72 bands corresponding to wavelengths from $367.7 nm$ to $1043.4 nm$ at a $9.5 - 9.6 nm$ spectral sampling interval. The spatial resolution is 1 m. \cite{gader:2013} provides more detailed information about the data. Two flights over the area from this data (Gulfport Campus Flight 1 and Gulfport Campus Flight 3) were selected as cross-validated training and testing data. Throughout the scene, there are 64 emplaced man-made targets shown in Fig. \ref{fig:gulfport_rgb}. The targets are cloth panels of four different colors: Brown (15 examples), Dark Green (15 examples), Faux Vineyard Green (FVG) (12 examples) and Pea Green (15 examples). We take Brown as our target type and for each target in the training flight, a $5\times5$ rectangular region around each ground truth point for each target were labeled as positive bags to account the drift coming from inaccurate GPS groundtruthing.

For the quantitative analysis of the estimated brown signatures, the normalized area of probability of detection (PD) under the curve (NAUC) at false alarm rate (FAR) $1\times 10^{-3} m^2$ were computed and were shown in Tab. \ref{tab:AUC_gulfport}, where the proposed MI-HE combined with HD preserves the best detection performance. Results reported are the median results over five runs.

%


	\section{Conclusion and Future Work}
	\label{sec:conclusion}
	
	The proposed MI-HE is able to learn target signatures with better quality and achieve competitive and state-of-the-art hyperspectral target detection results when compared to existing multiple instance concept learning methods.
The future work includes optimizing the objective function by quasi-newton method to improve its convergence and adding a discriminative term to promote the discriminativeness of the estimated target signature.
	
	\bibliographystyle{IEEEbib}
	\bibliography{MIHE_IGARSS_ref}
	
\end{document}